\documentclass[10pt, conference, compsocconf]{IEEEtran}
\IEEEoverridecommandlockouts

\usepackage{cite}
\usepackage{amsmath,amssymb,amsfonts}
\usepackage{algorithmic}
\usepackage{graphicx}
\usepackage{textcomp}
\usepackage{xcolor}

\def\BibTeX{{\rm B\kern-.05em{\sc i\kern-.025em b}\kern-.08em
    T\kern-.1667em\lower.7ex\hbox{E}\kern-.125emX}}
    
\begin{document}

\title{LingoMotion: An Interpretable and Unambiguous Symbolic Representation for Human Motion}


\author{\IEEEauthorblockN{Yao Zhang}
\IEEEauthorblockA{\textit{Aalto University} \\
Espoo, Finland \\
yao.1.zhang@aalto.fi}
\and
\IEEEauthorblockN{Zhuchenyang Liu}
\IEEEauthorblockA{\textit{Aalto University} \\
Espoo, Finland \\
zhuchenyang.liu@aalto.fi}
\and
\IEEEauthorblockN{Yu Xiao}
\IEEEauthorblockA{\textit{Aalto University} \\
Espoo, Finland \\
yu.xiao@aalto.fi}
}

\maketitle


\begin{abstract} 
Existing representations for human motion, such as MotionGPT, often operate as black-box latent vectors with limited interpretability and build on joint positions which can cause ambiguity. Inspired by the hierarchical structure of natural languages - from letters to words, phrases, and sentences - we propose LingoMotion, a motion language that facilitates interpretable and unambiguous symbolic representation for both simple and complex human motion.
In this paper, we introduce the concept design of LingoMotion, including the definitions of motion alphabet based on joint angles, the morphology for forming words and phrases to describe simple actions like walking and their attributes like speed and scale, as well as the syntax for describing more complex human activities with  sequences of words and phrases. The preliminary results, including the implementation and evaluation of motion alphabet using a large-scale motion dataset Motion-X, demonstrate the high fidelity of motion representation.
\end{abstract}
\begin{IEEEkeywords}
human activity recognition, human motion generation, human motion caption, interpretable representation, natural language, time-series
\end{IEEEkeywords}

\section{Introduction}
\label{sec:introduction}
Representing human motion is a foundational challenge in pervasive computing, spanning from human activity recognition (HAR) \cite{vrigkas2015review} to motion generation and captioning \cite{jiang2023motiongpt}. 
Regardless of the application, it is crucial to have an interpretable and unambiguous symbolic representation of human motion.


Previous joint position-based representations, such as MotionGPT \cite{jiang2023motiongpt}, VideoGPT \cite{yan2021videogpt} and ACTOR \cite{petrovich2021action}, encapsulate motion within black-box latent spaces with fixed temporal windows. These latent spaces are inherently opaque, offering limited interpretability and complicating the diagnosis or control of learned motion features. The fixed windowing process often disrupts the temporal continuity of sub-movements. Additionally, these holistic encodings lack compositional control, preventing users from flexibly assembling or editing motion using semantically meaningful components \cite{zhu2023human}. 




Recently, approaches such as PoseScript \cite{delmas2022posescript} and KinMo \cite{zhang2025kinmo} have adopted rule-based frameworks to generate textual descriptions for various body parts, rather than translating motions into a latent space. Although these methods improve interpretability, they are mainly suited for static poses and lack the integration of temporal information inherent to motion. Furthermore, the reliance on manually-crafted rules can result in distinct human poses being grouped under identical descriptions, causing potential ambiguity.

Meanwhile, the field of biomechanics has traditionally employed a decomposable, semantic approach to motion analysis, such as breaking down a gait cycle into distinct joint angle phases \cite{novacheck1998biomechanics}. This method allows complex motions, like running, to be divided into primary phases (e.g., Stance and Swing), each characterized by precise and coordinated movements of individual joint angles. Such an approach offers an interpretable representation that naturally incorporates the temporal dynamics of movement. Moreover, unlike joint-position-based representations, joint-angle-based analyses remain independent of global translation and orientation, providing a more robust framework for motion evaluation.

\begin{figure*}[ht]
    \centering
    \includegraphics[width=0.9\linewidth]{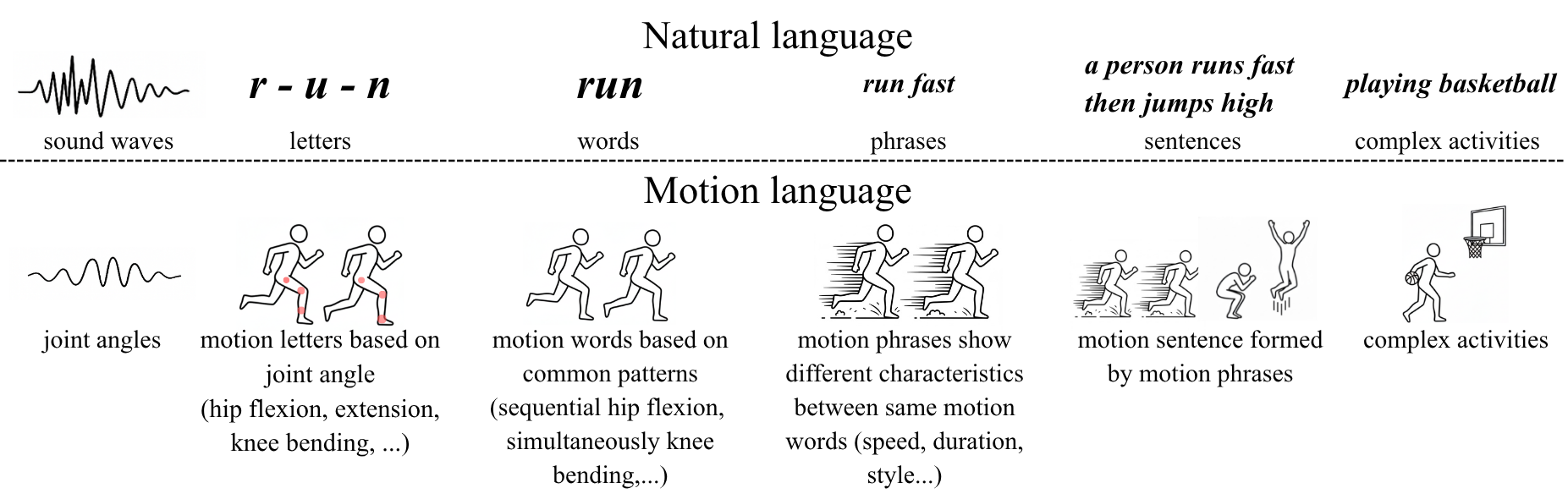}
    \caption{The core analogy of our paper: Just as natural language decomposes continuous sound waves into discrete letters, words, phrases, and sentences, we propose a "motion language" that decomposes continuous joint-angle signals into discrete motion letters, words, phrases, and sentences to describe complex human activities. This provides a structured, interpretable, and compositional framework for human motion representation.}
    \label{fig:compare_language}
\end{figure*}

This inspires us to propose LingoMotion, a novel motion language that facilitates interpretable and unambiguous symbolic representation of human motion based on joint angles. It adopts the hierarchical structure of natural languages. As illustrated in Fig. \ref{fig:compare_language}, 
we define motion "letters" based on joint angles, establish the morphology for forming "words" and "phrases" to describe simple actions like walking and their attributes such as speed and scale, and develop the syntax for articulating more complex human activities through sequences of words and phrases. This representation is highly interpretable; unlike "black-box" latent vectors, it is composed of "letters" that possess clear, biomechanically meaningful definitions. Because any motion can be decomposed into a unique sequence of these symbolic units, the representation is also unambiguous, paving the way for precise motion analysis, comparison, and reconstruction.

The key scientific contributions of this paper are summarized as follows. Firstly, we introduce LingoMotion, a novel motion language that facilitates interpretable and unambiguous symbolic representation of human motion, akin to the structure of natural language, laying a foundational framework for future tasks. Secondly, we present an initial segment-cluster-reconstruct pipeline designed to facilitate the discovery of a motion alphabet and the formation of motion words. Preliminary evaluations using Motion-X \cite{lin2023motion} illustrate the feasibility of reconstructing motion with high fidelity from motion letters and words, indicating a promising avenue for future research.


\section{Related Work}


Many recent works have focused on the challenges of decomposing motion into a series of sub-movements. For example, MotionGPT \cite{jiang2023motiongpt} and CoMo \cite{huang2024controllable} feed joint position based motion data through a fixed sliding window and embed it into a "black-box" latent space via Vector-Quantized Variational Autoencoders (VQ-VAEs) or generative transformers \cite{yan2021videogpt,petrovich2021action}. These latent vectors are intended to represent sub-movements, and results show that these embeddings indeed capture elements of the action. Similarly, Harish et al. \cite{haresamudram2024towards} explored this concept using IMU data, applying VQ-VAE to learn discrete representations for sub-movements. Their findings revealed consistent patterns across actions within the same class, suggesting that each action can be decomposed into a sequence of simpler movements. However, this implicit modeling approach has several drawbacks as discussed in Sec. \ref{sec:introduction}. To address these limitations, LingoMotion builds motion letters from joint angles, ensuring that each letter carries semantic meaning and remains interpretable, without cutting off sub-movements or losing information.




PoseScript \cite{delmas2022posescript} and KinMo \cite{zhang2025kinmo} have approached human motion with a focus on interpretability. They generate textual descriptions of poses based on a rule-based method, such as labeling the thigh–calf angle as "straight" if it exceeds 160°, or "partially bent" if it falls between 105° and 135°. While this produces a more interpretable pose descriptions, it is limited to static poses and does not incorporate the temporal information of the motion. Additionally, such descriptions can amalgamate distinct motions into a single category, leading to loss of details. In contrast, LingoMotion models motion as a dynamic sequence of joint-angle signals rather than static poses, allowing each motion to be represented as a unique composition of biomechanically meaningful elements.

\section{Concept Design}
\label{sec:motion Language concept}
By analogy with natural language, the motion language construct hierarchical motion letters, words, phrases, and sentences, thereby achieving an interpretable, symbolic, and unambiguous representation for human motion.

\subsection{Motion Alphabet}
\begin{figure}[t]
    \centering
    \includegraphics[width=1\linewidth]{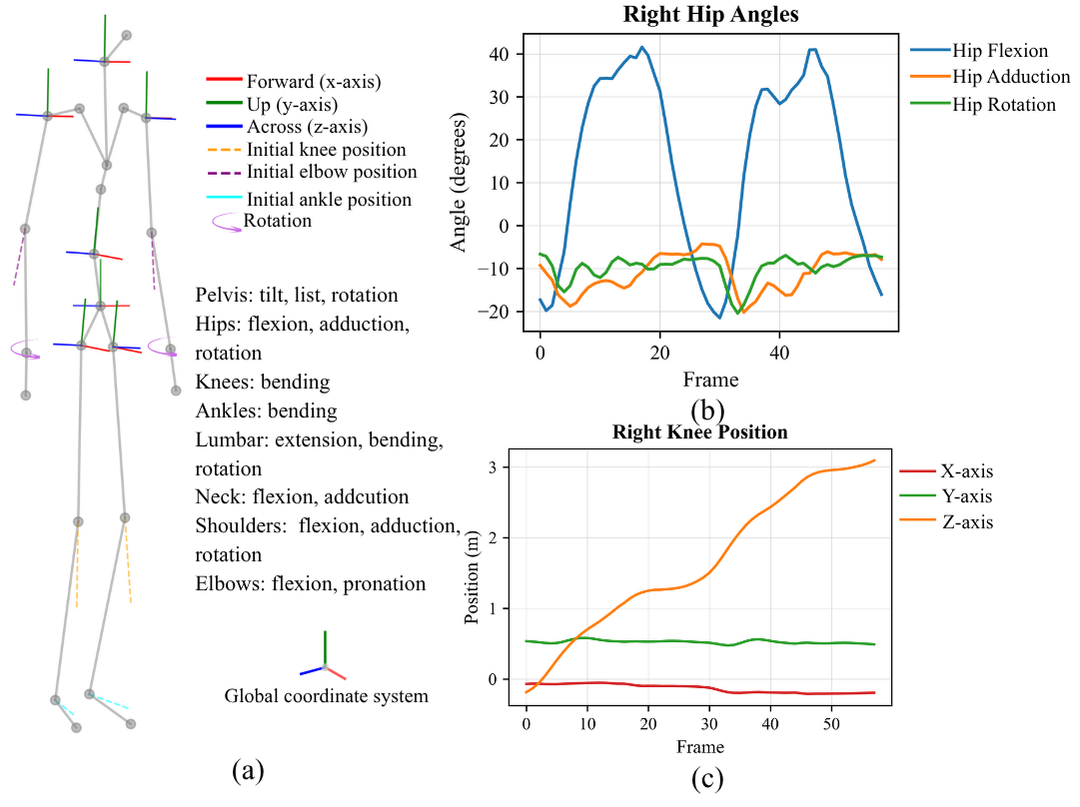}
    \caption{(a) Illustration of the 22 joints in the SMPL model, with joint angles calculated based on biological standards \cite{schlegel2024using, wu2002isb}. (b) Joint angles of the right hip, which has 3 DoFs (flexion, adduction, rotation), reveal periodic hip movement during walking. (c) The 3D positions of the right knee during walking, which lose the periodic information.}
    \label{fig:coordinate_representation}
\end{figure}

\begin{figure*}[htbp]
  \centering
  \includegraphics[width=0.9\linewidth]{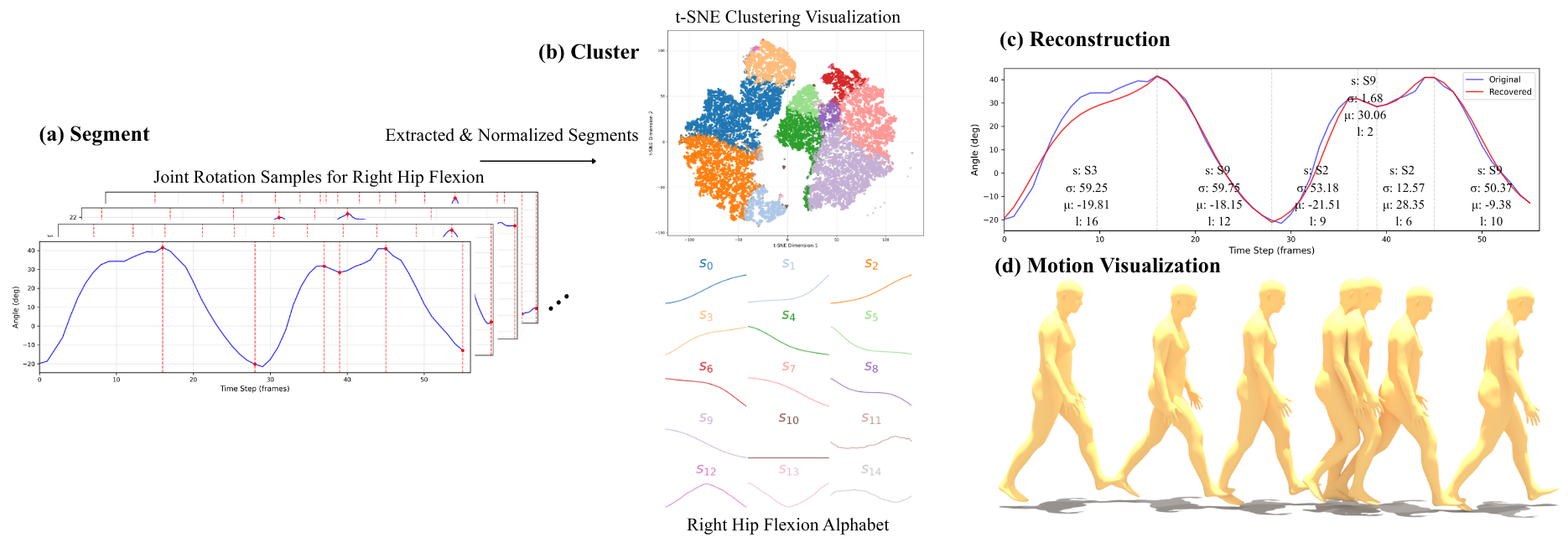}
  \caption{Our 3-stage pipeline for discovering the motion alphabet from right hip flexion measurements. (a) Segment the signal at local extrema. (b) Learn a codebook of canonical shapes via K-Means clustering on normalized segments. (c) Represent the original signal as a sequence of discrete "motion letters" (shape \textbf{s}) and their continuous attributes ($\sigma, \mu, l$). (d) Visualization of the 3D motion, rendered from the joint position, illustrating the corresponding physical poses at the segment boundaries.}
  \label{fig:coding_process}
\end{figure*}

The Skinned Multi-Person Linear model (SMPL) \cite{loper2023smpl} is a widely used parametric 3D human model for human motion representation. As illustrated in Fig. \ref{fig:coordinate_representation}(a), it models a human body with 22 important joints, with human motion being represented by the 3D position of these joints in the global coordinate system. Each joint has a different degree of freedom (DoF), due to the physiological structure of the human body. For instance, the right hip has 3 DoFs, including flexion, adduction and rotation, whereas the right knee only has one DoF, which is bending. All DoFs of a joint can be calculated as the angles between joint vectors and each plane of the local coordinate system. Detailed algorithms for joint angle calculation will be presented in Section IV.A. Consequently, the movement of a single joint with \(x\) DoFs can be described with \(x\) variables, each representing a joint angle.

We hypothesize that joint angles alone can provide interpretable and unambiguous representation of body poses and their changes along the time.  For example, during walking, the hip joint flexion from $-40^\circ$ to $60^\circ$, and its rotation changes between $-10^\circ$ and $-20^\circ$ to maintain balance. The positive or negative change in joint angle along each DoF has a clear physiological meaning \cite{novacheck1998biomechanics}. For example, a hip flexion could be described as the angle between the thigh vector and the ZY plane in its local coordinate system ranging from -40 to 40 degrees, while the angles with the other two planes remain unchanged. We segment the motion into a sequence of such motion letters, as described in IV.A.

There are two primary reasons for choosing joint angles over joint positions. Firstly, joint positions often fail to capture essential features such as periodic repetition. For instance, as illustrated in Fig. \ref{fig:coordinate_representation}(b) and (c), the periodic movement of the hip joint is not clearly represented through joint positions, while joint angles effectively capture this information. Secondly, joint positions can be influenced by variations in body shape, whereas joint angles remain robust to such differences.

\subsection{Motion words, phrases and sentences}

    \textbf{Motion Words:} The coordination of
    various joints is integral to human movement \cite{novacheck1998biomechanics}. For
    example, during walking, one leg flexes from $-20^\circ$ to
    $20^\circ$, while the other leg flexes from $-20^\circ$ to $20^\circ$, and
    simultaneously, the knee bends from $10^\circ$ to $30^\circ$ \cite{kraus2005comparative}.
    We propose to form motion words from relevant motion letters, taking into account the temporal relationship (e.g., simultaneously, sequentially).
    
    \textbf{Motion Phrases:} Joint movements exhibit varying speeds, durations, and amplitudes, which can not be fully captured by motion letters alone. Inspired by the shapelet concept \cite{ye2009time}, we represent each constituent letter as a four-element tuple, comprising the letter and its attributes: $ ( \text{scale}, \text{bias}, \text{length})$, as shown in Fig. \ref{fig:coding_process}. The length determines the duration of the letter, the scale determines the amplitude, and the bias ensures continuity between letters. This allows us to form different motion phrases (e.g., "walk fast" vs. "walk slowly") by adjusting these attributes.
    
    \textbf{Motion Sentences:} Motion words and phrases can be combined to form sentences, representing complex actions like "a person runs fast then jumps high" or "playing basketball". These combinations, or "specific patterns", are primarily defined by their temporal relationship. In the simplest scenario, this involves a simple sequential order. For more complex activities (like "playing basketball"), this "syntax" can be modeled using advanced sequential models, such as Transformers\cite{vaswani2017attention}, which can learn the probabilistic patterns of how motion words and phrases co-occur and transition over time within the symbolic sequences.

\section{Preliminary Implementation and Evaluation}
\label{sec:preliminary results}

\subsection{Discovery of Motion Alphabet}

Taking the right hip flexion as example, Fig. \ref{fig:coding_process} illustrates a 3-stage pipeline that transforms continuous joint angle measurements into a sequence of discrete motion letters. 
First, it decomposes continuous motion signals into a series of motion letters based on detecting transition between positive and negative changes. Next, the pipeline clusters these primitives based on their shape characteristics, with each cluster center defining a single "motion letter", thereby forming an "alphabet" composed of standard motion patterns. 
Each letter in the sequence is then associated with attributes calculated directly from the statistics of the original motion segment, including amplitude as "scale", minimum value as "bias", and duration as "length". This association ensures that the discrete representation can accurately reconstruct the details of the original motion.

\subsection{Conversion from joint positions to joint angles}
Given joint positions as 3D coordinates within a global coordinate system, joint angles can be calculated from joint positions in four steps, according to established biomechanical standards\cite{wu2002isb}. Here the global coordinate system refers to the fixed, external reference coordinate. First, the local coordinate system (LCS) for a given joint is kinematically defined relative to the joint angle of its physiologically proximal segment. For example, the pelvis angle is derived from the orientation of the plane formed by the pelvis and the hip centers with respect to the global coordinate system planes. Next, the hip's LCS is established by rotating the global coordinate system based on the calculated pelvis angle. The hip's 3-DoF joint angle is then determined by the angles between the thigh vector and the planes of the hip LCS. The knee's LCS is subsequently derived by rotating the hip LCS based on the determined hip joint angles. 

We calculated the joint angles from joint positions for all the motion sequences in the motion-x dataset, which comprises 81.1K motion sequences. After that, we recalculate the joint positions from these joint angles, and compare them with the original joint positions. As shown in Table \ref{tab:recon_errors}, the joint positions can be recovered with high accuracy, demonstrating the fidelity of the conversion from joint positions to angles.

\subsection{Fidelity of motion representation}

We evaluate the fidelity of  motion reconstructed from a sequence of motion letters. For this evaluation, we divide the motion-x dataset into a training set (70\%) and a test set (30\%). The first two stages of our pipeline (segmentation and clustering) are run on the training set to learn the alphabet codebooks. In this preliminary implementation, we define the alphabet across 28 joint angle channels. Since each channel's movement patterns are clustered independently, our "motion alphabet" consists of 28 separate codebooks. The number of "letters" in each codebook varies based on the motion's complexity, in our experiment ranging from 12 (e.g., for right hip rotation) to 20 (e.g., for right elbow flexion).

We encode the unseen test set into discrete $(\mathbf{s}, \sigma, \mu, l)$ representation based on learned codebooks and then reconstruct the joint-angle signals. The error is calculated by comparing these reconstructed joint-angle signals against the original ground truth angle signals from the test set. As shown in Table \ref{tab:recon_errors} and Fig. 3(c), this symbolic representation still retains a high degree of fidelity. 




\begin{table}[tbp] 
\caption{Accuracy of representation learning. RMSE (Root Mean Square Error) and NMSE (Normalized Mean Square Error) are reported.}
\label{tab:recon_errors}
\begin{center}
\begin{tabular}{@{}l c c c@{}} 
\hline
\textbf{Experiment} & \textbf{RMSE} & \textbf{NMSE} & \textbf{R²} \\
\hline 
Joint positions $\rightarrow$ angles $\rightarrow$ positions & 0.009 & 0.005 & 0.978 \\
Joint angles $\rightarrow$ motion letters $\rightarrow$ joint angles & 0.921 & 0.103 & 0.716 \\
\hline
\end{tabular}
\end{center}
\end{table}

\section{Conclusions and Future Work}
We have introduced the concept design of LingoMotion and the preliminary implementation and evaluation of the Motion Alphabet, the initial component of LingoMotion. The early results are promising, demonstrating high-fidelity reconstruction from a set of interpretable, biomechanically meaningful motion letters. Our future work will focus on expanding the implementation to support the formulation of phrases and sentences, as well as addressing remaining challenges. Firstly, we aim to ensure that the learned cluster centers genuinely represent the most meaningful motion letters indicative of typical movements. Secondly, we will explore the convergence of the motion alphabet to verify whether the number of letters stabilizes as the dataset expands. Thirdly, we need to develop models that can effectively handle the variable-length sequences of motion words. Additionally, we'll investigate how to leverage "motion language" to align motion with natural language, uncovering common "syntax" or co-occurrence patterns between motion letters, words, and phrases.



 \begingroup
\bibliographystyle{IEEEbib}
\bibliography{ref}
\endgroup

\end{document}